\documentclass{article}
\usepackage{ijcai20}

\usepackage{times}
\usepackage{soul}
\usepackage{url}
\usepackage[hidelinks]{hyperref}
\usepackage[utf8]{inputenc}
\usepackage[small]{caption}
\usepackage{graphicx}
\usepackage{amsmath}
\usepackage{amsfonts}
\usepackage{amsthm}
\usepackage{algorithm}
\usepackage{algorithmic}
\usepackage{booktabs} 
\usepackage{todonotes}
\usepackage{subcaption}
\usepackage{multirow}
\usepackage{enumitem}
\usepackage{comment}

\title{Robust Deep Neural Networks Inspired by Fuzzy Logic}

\author{
    Minh Le
    \affiliations
    Vrije Universiteit Amsterdam, Netherlands
    \emails
    m.n.le@vu.nl
}

\begin{document}

\maketitle

\begin{abstract}
Deep neural networks have achieved impressive performance and become the de-facto standard in many tasks.
However, troubling phenomena such as adversarial and fooling examples suggest that the generalization they make is flawed.
I argue that among the roots of the phenomena are two geometric properties of common deep learning architectures: their distributed nature and the connectedness of their decision regions. 
As a remedy, I propose new architectures inspired by fuzzy logic that combine several alternative design elements.
Through experiments on MNIST and CIFAR-10, the new models are shown to be more local, better at rejecting noise samples, and more robust against adversarial examples.
Ablation analyses reveal behaviors on adversarial examples that cannot be explained by the linearity hypothesis but are consistent with the hypothesis that logic-inspired traits create more robust models.
\end{abstract}

\section{Introduction}
\label{sec:intro}

The success of ReLU-based neural networks in recent years originates from their \textit{learnability} and \textit{expressiveness}. 
Piecewise linearity helps them avoid gradient exploding or vanishing problems of early architectures and, in terms of expressive power, they are shown to be able to capture an exponential amount of variations \cite{Raghu2017}. 
However, little can be guaranteed about the \textit{validity} of the learned representations.
Recent work has pointed out two troubling phenomena: adversarial examples \cite{Szegedy2013} and fooling examples \cite{nguyen15}. 
In the context of image recognition, the former are images slightly perturbed to fool a model while being semantically the same to human eyes. The latter, fooling examples, are images that do not belong to any class and yet are classified to one with high confidence.
Both the results of Szegedy et al.~\shortcite{Szegedy2013} and Nguyen et al.~\shortcite{nguyen15} demonstrate that a model that is easy to train might also easily make invalid generalization.


Much research has been devoted to explaining these phenomena \cite[among others]{Goodfellow2015,Ford2019}.
Recently, Nguyen et al.~\shortcite{Nguyen2018} identifies the tendency of ReLU networks to create connected classification regions as one likely source of adversarial examples.

With regards to fooling examples, the networks' distributed nature is likely to be among the causes. The activation of a ReLU unit gets stronger the \textit{further} the input is from its decision boundary.
Noise patterns are typically far from decision boundaries and therefore also likely to induce a high level of activation.
A more principled approach would be to make \textit{local} generalizations in which a model makes confident predictions within bounded regions of the input space and gradually becomes less confident further away.
Even though removing the locality assumption is among motivations to develop deep architectures \cite{bengio2011expressive}, it will be shown here that we can have one that is both deep and local.

To design a local and disconnected deep neural network, I take inspiration from fuzzy logic.
Image classification can be cast as learning a system of propositions that take pixel intensities as input and produce the likelihood of each class being present.
Given the complexity of the task, it is reasonable to expect those propositions to be expressed in terms of concepts which, in turn, are expressed by sub-concepts until we reach the input, for example:
\[
\begin{array}{lcl}
\mathrm{class}_1 & \leftarrow & \mathrm{concept}_1 \lor \mathrm{concept}_2 \\
\mathrm{class}_2 & \leftarrow & (\mathrm{concept}_2 \land \mathrm{concept}_4) \lor \mathrm{concept}_3 \\
\cdots \\
\mathrm{concept}_n & \leftarrow & (x_{0,0} \land x_{0,1}) \lor (\neg x_{0,1} \land x_{1,1} \land x_{1,2}) \\
\end{array}
\]

The hierarchical layout and alternating pattern of \textsc{AND}s and \textsc{OR}s above are essential for expressiveness because they compactly express an exponential number of cases.
Concepts, as specified by fuzzy propositions, can be local and disconnected in the sense that they admit a typically small subset of the input space (e.g.\ dogs among all visible objects) and can combine arbitrary subconcepts into one (e.g.\ to account for the diverse appearances of birds, from robins to ostriches to penguins).
It will be shown in next sections that it is possible to incorporate such abilities into a modern neural network architecture and doing so helps alleviate fooling and adversarial examples.

The rest of paper is organized as follows: modifications to a ReLU network is proposed in Section~\ref{sec:logic-neural}; experimental procedures and results are detailed in Section~\ref{sec:experiments} and \ref{sec:adversarial};
I will situate the research in the wider literature in Section~\ref{sec:related-work} before concluding with some remarks in Section~\ref{sec:conclusions}.

\section{Approximating Propositional Logic in Neural Networks}
\label{sec:logic-neural}

The task of emulating logic in a high-dimensional space carries with it inherent difficulty. To balance learnability and capacity, I start with ReLU-based CNNs and make amendments where they deviate from the ideals. The results are captured in this section as a series of re-considerations.

\subsection{Rethinking Activation Functions}
\label{sec:relog}
If we are to think of a neuron as representing a logic formula, constraints must be placed on its range of activation.
Common sense suggests that a soft logic function ramps up its activation as more and more supporting evidence accumulates, but only to a certain point. In other words, it must exhibit a saturation region. 
Sigmoid achieves this by exponential decay but this has been shown to cause vanishing gradient problem.

\begin{figure}
    \centering
    \small
    \newcommand{\subwidth}{0.23\textwidth} 
    \begin{subfigure}[b]{\subwidth}
        \includegraphics[width=\textwidth]{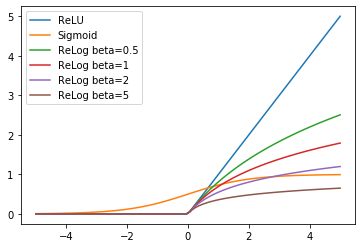}
        \caption{Activation}
        \label{fig:activation-function}
    \end{subfigure}
    \begin{subfigure}[b]{\subwidth}
        \includegraphics[width=\textwidth]{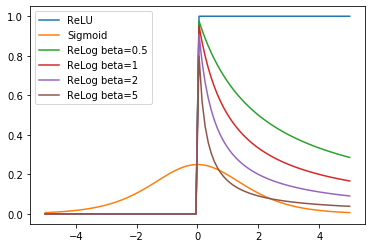}
        \caption{Derivative}
        \label{fig:activation-function-derivative}
    \end{subfigure}
    \caption{Comparing ReLog with two popular activation functions}
\end{figure}

To strike a balance between fidelity and learnability, I introduce rectified logarithmic function (Figure~\ref{fig:activation-function}):

\begin{equation}
\textrm{ReLog}(z; \beta) = \frac{1}{\beta} \log \left( \beta \max(z, 0) + 1 \right),
\end{equation}
where $\beta > 0$.

As can be seen in Figure~\ref{fig:activation-function-derivative}, in the right half of the function, the derivative of ReLog peaks higher and stays meaningful for a much larger range compared to sigmoid.
Another important property of ReLog is that it approximates ReLU when $\beta$ is close to 0.

{}

\subsection{Rethinking Neurons}
\label{sec:minmax-quadratic}

In the distributed representations framework, the individual neuron does not carry any particular meaning \cite{Hinton86}. In contrast, a logical view considers the neuron as a function that returns true for a particular pattern and false for everything else. 

\begin{figure}
    \centering
    \begin{subfigure}[b]{0.15\textwidth}
        \includegraphics[width=\textwidth]{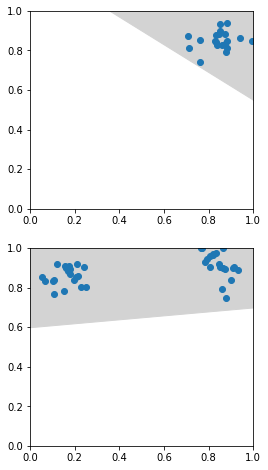}
        \caption{ReLU\\\hspace{\textwidth}}
        \label{fig:filter-relu}
    \end{subfigure}   
    ~ \begin{subfigure}[b]{0.15\textwidth}
        \includegraphics[width=\textwidth]{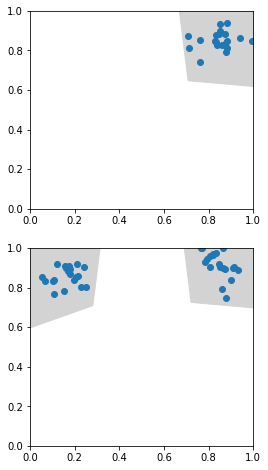}
        \caption{Min-max-out\\\hspace{\textwidth}}
        \label{fig:filter-maxout}
    \end{subfigure}   
    ~ \begin{subfigure}[b]{0.15\textwidth}
        \includegraphics[width=\textwidth]{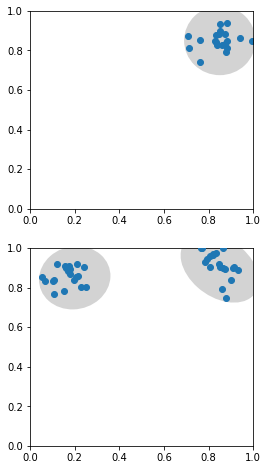}
        \caption{Elliptical kernel + Max-out}
        \label{fig:filter-quadratic-maxout}
    \end{subfigure}       
    \caption{Different types of neurons differ in the ability to fit to the shape of the data. Blue dots represent data points and shaded regions are regions in the input space that creates positive activation in the target neuron.}
    \label{fig:filters}
\end{figure}

The patterns depicted in each column of Figure~\ref{fig:filters} can be captured by logical conjunctions and disjunctions which translate into $\min$ and $\max$ in fuzzy logic \cite{Belohlavek2018}. For this reason, I propose min-max-out units, an extension of maxout \cite{goodfellow2013maxout.manual} in which incoming input is first passed through a $\min$ operation (representing \textsc{AND}) and then a $\max$ operation (representing \textsc{OR}):
\begin{equation}
    \textrm{MinMaxOut}_{m,n}(x) = \max_{i=1}^m \min_{j=1}^{n} \sum_{k=1}^p w_{ijk} x_k
\end{equation}

The result is a better fit as illustrated in Figure~\ref{fig:filter-maxout}. However, because the number of linear units needed to fit a point cloud increases linearly with the number of dimensions, this solution might not scale to high-dimensional spaces. 
An alternative is to draw curved boundaries around the data points by applying the kernel trick:

\begin{equation}
    z = \sum_{i=1}^m w^\prime_i x_i^2 + \sum_{i=1}^m w_i x_i + b,
\label{eq:ellipse}
\end{equation}

To create elliptical boundaries, the constraint $w^\prime_i \le 0\ \forall i$ can be applied.
Elliptical units eliminate the need for min-out but not max-out units as illustrated in Figure~\ref{fig:filter-quadratic-maxout}.

{}

\subsection{Rethinking Regularization}

Explicit regularization is a way to reduce overfitting. This is clear in the textbook case of fitting a polynomial curve where smaller weights can reign in curvature \cite{Bishop2006}. 
However, it is not obvious if and how this applies to a ReLU-based neural network that is linear almost everywhere. 
The analysis in the previous section suggests an alternative view on regularization as margin modulation. 

The distance between a ReLU unit's decision boundary and a training example in Cartesian space is:
\begin{equation}
    d = \frac{\left| \sum_{i=1}^{m} w_i x_i + b \right|}{\sqrt{\sum_{i=1}^{m} w_i^2}},
\end{equation}
where $x \in \mathbb{R}^m$ is the input vector, $w \in \mathbb{R}^m$ and $b\in \mathbb{R}$ are the weights and bias of the neuron.

Similar to a support vector machines (SVM), the distance between the decision boundary and the closest example can be increased by minimizing $||w||$ \cite{Bishop2006}. Notice that, different from SVM, only accepted patterns count because rejected ones lead to inactivation and zero gradient.

On the other hand, the decision boundary is removed from the coordinate origin by:
\begin{equation}
d_0 = \frac{|b|}{\sqrt{\sum_{i=1}^{m} w_i^2}}.
\end{equation}

Noticing that the origin is on the same side as the accepted input pattern if $b>0$ and on the opposite side if $b<0$, it can be easily shown that reducing $b$ reduces the margin around the accepted pattern.
Combining the two observations above gives us a way to control decision margin that I shall call max-fit for its ability to increase fitting to an input pattern: 
\begin{equation}
    r_\text{max-fit} = \frac{\gamma}{m} \sum_{i=1}^{m} |w_i| + \gamma^\prime b 
\end{equation}
where $\gamma, \gamma^\prime \in \mathbb{R}^+$ govern the strength of regularization and $m$ is the number of inputs.
A similar formula can be defined for $L_2$ regularization.

    

\subsection{Rethinking Training}
\label{sec:bce}


The approximation would not be complete without the ability to reject examples that do not belong to any of the predefined classes (i.e.\ to compute a all-zero row in a truth table).
To this end, I replace softmax with independent sigmoid activation and train directly on negative examples using binary cross-entropy (BCE) loss. 
I also experimented with mean square loss but found that it leads to slower convergence and lower accuracy.
Because techniques to generate fooling examples are out of the scope of the current paper, I only experimented with overlaying two images from different classes. 
The images can be efficiently generated from training examples as a data augmentation step.

\section{Experimental Design}
\label{sec:experiments}

Experiments were designed to validate the following hypotheses about a model that contains some of the proposed modifications:
\begin{description}[noitemsep,nolistsep]
\item[H1 (learnability and expressiveness):] the model can be trained to the same accuracy on a clean dataset.
\item[H2 (locality):] for natural examples, a class is predicted with high confidence which decreases as we go further away until all classes are assigned equal probabilities.
\item[H3 (robustness):] the model achieves a higher accuracy on adversarial examples compared to a ReLU network.
\end{description}

It is straightforward to compare models on clean and adversarial examples. 
To detect locality, I use a linear interpolation between a natural image and a noise pattern. 
The expectation is that a distributed model is strongly activated everywhere along the line while a local model is only strongly activated in the vicinity of the image.
I also perform ablation analyses to detect the benefit of each proposed trait.

\subsection{Datasets}

\textbf{MNIST} \cite{Lecun1998} is a widely used dataset in image recognition. It contains $28 \times 28$ gray-scale images of hand-written Arabic digits, divided into 60,000 examples for training and 10,000 for testing. The small scale enables fast experimentation and therefore it is commonly used to evaluate proofs of concept.

\textbf{CIFAR-10} \cite{krizhevsky09} contains slightly larger ($32 \times 32$) color images of 10 classes of objects in natural settings. There are 50,000 training and 10,000 testing examples. Compared to the previous dataset, CIFAR-10 is harder in both natural and adversarial settings.

\subsection{Models}

For MNIST, I experimented with a simple model with two convolutional layers (16 and 32 channels, both with $5 \times 5$ kernels) and a densely connected output layer.
This model and its derivatives are trained using Adam with learning rate 0.001 for 20 epochs.
Data augmentation methods used include resizing, erasing, and rotating randomly, and adding Gaussian noise ($\sigma=0.3$).

For CIFAR-10, I used a scaled-down version of VGG architecture \cite{Simonyan.zisserman2014} with 8 convolutional layers. This choice helps to illustrate how a deeper architecture works while being lightweight enough for quick experimentation.
All models in this line are trained using mini-batch SGD with learning rate 0.01 for 80 epochs. 
For data augmentation, I used random cropping, horizontal flipping, rotation, and Gaussian noise ($\sigma=0.5$, added after normalization).

In both experiments, I start with a baseline model that uses a ReLU architecture and is trained on only natural examples using cross entropy loss. Modifications are added one by one to observe the individual effect (except for elliptical units where I also remove min-out because it becomes unnecessary while making training unstable).

\subsection{Training with Quadratic and Logarithmic Units}
\label{sec:training}

The piecewise linearity of ReLU networks creates meaningful gradient almost everywhere therefore enables efficient training. In contrast, architectures with highly non-linear elements are very hard to train.
I observe that vanilla implementations of logarithmic and elliptical nets tend to train poorly: the performance oscillates around chance level early on from the first epochs or decrease substantially midway through.

To achieve training stability, I start by training a ReLU network and gradually ramp up nonlinear elements.
I also find it beneficial to detect collapse events, defined as a decrease of training accuracy by a half in an epoch. In such cases, the model is recovered to a checkpoint and trained for 5 more epochs without increasing nonlinearity.

One might notice that linear functions are simply quadratic ones with second-degree coefficients set to zero. Therefore, to train quadratic units (Section~\ref{sec:minmax-quadratic}), I replace Equation~\ref{eq:ellipse} with:
\begin{equation}
    z = \alpha_t \sum_{i=1}^m w^\prime_i x_i^2 + \alpha_t \gamma + \sum_{i=1}^m w_i x_i + b,
\end{equation}
where $t$ is training step, $a_t \in [0,1]$, and $\gamma > 0$. 
The term $\alpha_t \gamma$ is added to the net input to compensate for a possible increase in negativity coming from quadratic terms. Because gradient only flows through active neurons, gaining new active units is a smaller concern than losing existing ones.
For a network initialized with Kaiming method, I find that $\gamma = 1$ is sufficient to stabilize learning.

Similarly, ReLog activation function (Section~\ref{sec:relog}) is a generalization of ReLU in the sense that the former approaches the latter when $\beta$ tends to 0. Therefore, I compute the activation of a neuron as:
\begin{equation}
    y = \mathrm{ReLog}(z;\alpha_t^\prime \beta)
\end{equation}
where $z$ is the net input and $\alpha_t^\prime$ grows from 0 to 1 with each training step.

Many hyperparameters are involved in the various models. For practical reasons, I only tried a few combinations that make intuitive sense and select the best one with respect to performance on natural datasets.

\subsection{Attacks}

The importance of evaluating against strong attacks has been highlighted more than once in the literature \cite[for example]{Athalye2018.short,Uesato2018.short}. I used Cleverhans
, a library of standard implementations of state-of-the-art attacks. Among the algorithms offered there, I selected a battery of attacks from different categories: single-step: \textbf{FGM} \cite{Goodfellow2015}; iterative: \textbf{BIM} \cite{Kurakin2017} and \textbf{C\&W} \cite{Carlini2017}; and gradient-free: \textbf{SPSA} \cite{Uesato2018.short}.

For each combination of attack, model, and dataset, I evaluated on batches of 100 test examples, using between 15 and 100 batches depending on the speed of the attack (SPSA is the slowest, taking more than 3 hours to finish a batch). The accuracy on batches is used to calculate average performance and $p$-value.\footnote{I use the function \texttt{ttest\_ind} in SciPy package with \texttt{equal\_var=False}.}

\section{Results}
\label{sec:adversarial}

In this section, I will present the results of the proposed experiments and explain how they validate my hypotheses.

\subsection{Learnability and Expressiveness}
\label{sec:learnability}

The performance of my models on clean MNIST digits can be found in Table~\ref{tab:noise-classification}.
It is clear that we can reach similar levels of performance with the proposed architecture under a conventional training regime.
Training with negative examples hurts the performance, especially on CIFAR-10.
A possible explanation is a propositional theory of CIFAR-10 might be excessively big and a more expressive form of logic might be needed.
Alternatively, a neural network might need higher capacity to represent such a propositional theory.

The ``Proposed arch.''\ results show that we can train a modified architecture with elliptical units and ReLog activation function to the same or better performance compared to ReLU networks.
To confirm that filters are actually quadratic, I examine the magnitude of quadratic weights.
For an MNIST model, the median of quadratic weights is 1.75 and 10.76 times that of linear weights for the first and the second quadratic layers, respectively.
For the CIFAR-10 model, the corresponding figures fall between 0.33 and 1.33.

\subsection{Locality}
\label{sec:locality}

\newcommand{\digittonoise}[1]{
    \begin{subfigure}[b]{0.35\textwidth}
        \includegraphics[width=\textwidth]{images/transition-num#1.png}
    \end{subfigure}   
}
\newcommand{\dnactivation}[1]{
    \begin{subfigure}[b]{0.35\textwidth}
        \includegraphics[width=\textwidth]{images/predictions-num#1.png}
    \end{subfigure}
}
\newcommand{\dvsl}[1]{
    \digittonoise{#1}
    
    \dnactivation{#1}
}
\begin{figure}[t]
    \centering
    \dvsl{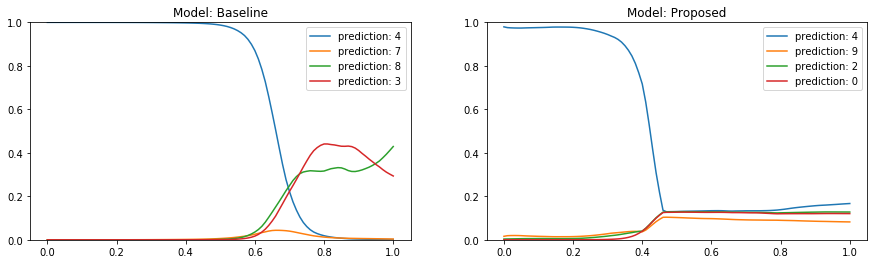}
    \dvsl{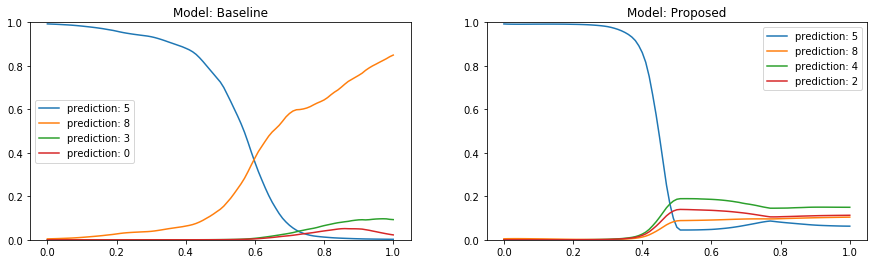}
    \dvsl{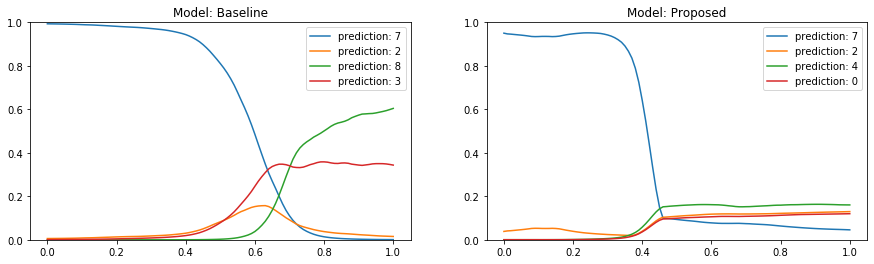}
    \dvsl{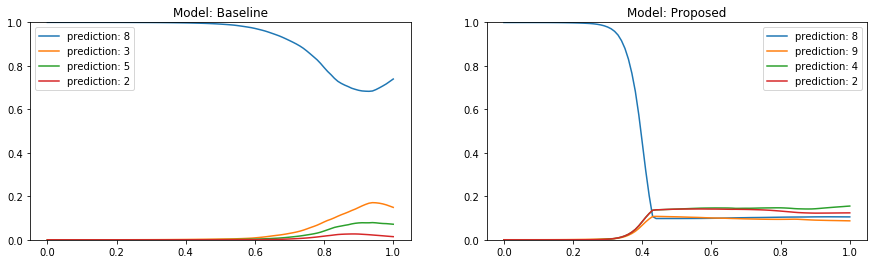}
    \caption{Comparing activation patterns of ReLU and the proposed architecture on noisy images. Horizontal axis: level of noise, vertical axis: softmax activation.}
    \label{fig:representations}
\end{figure}{}

\begin{table}
    \small
    \centering
    \begin{tabular}{lcccc}
\toprule
\multirow{2}{*}{Model}  & \multicolumn{2}{c}{Real images} & \multicolumn{2}{c}{Noise} \\
& Acc. & Prob. &  Accuracy &  Probability \\
\midrule
\multicolumn{5}{c}{\hspace{1em}MNIST} \\
\midrule
Baseline &         0.99 & 0.99 &    $0.22 \pm 0.15$ & 0.64 $\pm$ 0.08 \\
Proposed arch. &         0.99 & 0.98 &    $0.81 \pm 0.35$ & $0.35 \pm 0.19$ \\
Neg.\ training &         \textbf{1.00} & 1.00 &     0.27 $\pm$ 0.13 & 0.64 $\pm$ 0.09 \\
All mod.  &         0.95 &  0.95 &    $\textbf{0.92} \pm 0.25$ & $0.21 \pm 0.16$ \\
\midrule
\multicolumn{5}{c}{\hspace{1em}CIFAR-10} \\
\midrule
Baseline & 0.75 & 0.91 & 0.00 & 1.00 \\
Proposed arch. & \textbf{0.77} & 0.84 & 0.05 & 0.59  \\
All mod. & 0.63 & 0.75 & \textbf{0.17} & 0.64 \\
\bottomrule
\end{tabular}
    \caption{Performance on the task of classifying real images and noise patterns. Models: Baseline = ReLU, Proposed arch.\ = ReLog + Elliptical + MaxOut + MaxFit ($L_1$), Neg.\ training = BCE training + Negative examples, All mod.\ = Proposed architecture + Negative training. Performance on MNIST noise is reported as mean $\pm$ std as measured on 10 models trained with different random seeds.}
    \label{tab:noise-classification}
\end{table}{}

Figure~\ref{fig:representations} depicts the activation pattern of the baseline model and the model with all logic-inspired traits enabled.
I plotted the softmax-normalized predictions of four classes that are selected to include the highest activation for either the given example or the noise pattern.

It is abundantly clear that, whereas the ReLU model emits high probabilities for both true digits and noise, the proposed model is only activated around the test instances.
Hyperparameters can argurably be tuned to make the model more robust to noise but this is beyond the scope of the paper.

To evaluate the behavior quantitatively, I obtain models' predictions on 1,000 noise images (sampled uniformly in [0,1] for MNIST and from $\mathcal{N}(0,1)$ for CIFAR-10 because, for the latter, pixel intensity is normalized during preprocessing). Any prediction with higher than 50\% confidence is labeled as an error.
As I notice that the performance on MNIST noise pattern shows high variance, I trained and evaluated 10 models using different random seeds.

Table~\ref{tab:noise-classification} demonstrates the high efficiency of the proposed models on rejecting noise for MNIST. The difference in predicted probability for real images and noise indicates that the models are local.
Comparing the performance of the proposed architecture and negative training on a ReLU architecture, one can conclude that while negative training helps improves noise rejection, the new architecture is essential to reach high performance.

On CIFAR-10, the new architecture increases the chance of rejecting noise and reduce the prediction confidence compared to the baseline. 
There is also a pronounced difference between probabilities predicted by ``Proposed arch.'' on real images and on noise patterns, suggesting that the model is more local than a ReLU network.
The negatively trained model is better at rejecting noise at the cost of a lower accuracy on true images.

\subsection{Robustness on MNIST}
\label{sec:mnist-results}

\newcommand\s{\hspace{0.5em}}
\begin{table*}[t]
    \centering
    \small
\begin{tabular}{rlcccccc}
\toprule
 & \multirow{2}{*}{Model} & Clean & FGM & FGM & C\&W & BIM & SPSA \\
&  & & $\epsilon_{L_\infty}=0.3$ & $\epsilon_{L_2}=2$ & & $\epsilon=0.3$ & $\epsilon=0.3$ \\
\midrule
0 &             Baseline &         \s 0.99\hspace{0.5em} &            \s 0.07\hspace{0.5em} &           \s 0.69\hspace{0.5em} &       \s 0.01\hspace{0.5em} &        \s 0.00\hspace{0.5em} &         \s 0.07\hspace{0.5em} \\
1 &     + ReLog ($\beta=2$) &         \s 0.99$^*$ &            \s 0.30$^*$ &           \s 0.84$^*$ &       \s 0.19$^*$ &        \s 0.01$^*$ &         \s 0.37$^*$ \\
2 &       + MaxOut ($k=4$) &         \s 0.99$^*$ &            \s 0.42$^*$ &           \s 0.86$^*$ &       \s 0.23$^*$ &        \s 0.03$^*$ &         \s 0.44$^*$ \\
3 &       + MinOut ($k=2$) &         \s 0.99\hspace{0.5em} &            \s \textbf{0.66}$^*$ & \s \textbf{0.92}$^*$ &       \s 0.31$^*$ & \s \textbf{0.16}$^*$ &         \s \textbf{0.52}$^*$ \\
4 &         + Elliptical ($\alpha=1$) &         \s 0.93$^*$ & \s 0.58$^*$ & \s 0.57$^*$ &       \s \textbf{0.47}$^*$ &        \s 0.02$^*$ &         \s 0.22$^*$ \\
5 &        + MaxFit ($L_1$) & \s 0.99$^*$ & \s 0.35$^*$ & \s 0.82$^*$ & \s 0.19$^*$ &        \s 0.02\hspace{0.5em} &         \s 0.35$^*$ \\
6 &       + BCE training &        \s  0.98$^*$ &            \s 0.56$^*$ &          \s  0.83$^*$ &       \s 0.26$^*$ &        \s \textbf{0.16}$^*$ &         \s 0.42$^*$ \\
7 &  + Negative examples &         \s 0.96$^*$ &            \s 0.24$^*$ &           \s 0.66$^*$ &       \s 0.12$^*$ &        \s 0.00$^*$ &         \s 0.14$^*$ \\
\bottomrule
\end{tabular}
    
    \caption{Accuracy of models on MNIST on different attacks. Elliptical units eliminate the need of min-out so I disable it from step 4 onward. BIM: iterations=5, C\&W: iterations=50, SPSA: iteration=50. An asterisk ($^*$) signifies that the result is statistically significantly different from the previous one on the same column.}
    \label{tab:accuracy}
\end{table*}

Table~\ref{tab:accuracy} presents the robustness of my models against various attacks. It is clear that the proposed architectures lead to a significant increase in robustness across many types of threats.
Four out of six models (Row~2, 3, 5, and 6) are more robust than the baseline on all measures.
The remaining two models beat the baseline on all but one attack.

Since all models are trained on standard back-propagation, it is unlikely that the improvements come from gradient obfuscation \cite{Athalye2018.short}. There are other signs that this is not the case: single-step attacks (variants of FGM) and gradient-free attack (SPSA) are generally not more effective than iterative gradient-based attacks (C\&W and BIM).

One of the leading hypotheses about the origins of adversarial examples states that it is the local linearity of ReLU models that makes attacks possible \cite{Goodfellow2015,Warde-farley2018}.
This hypothesis is compatible with the observation that ReLog is more robust than ReLU against adversarial perturbation.
However, it is contradicted by the observation that elliptical units, despite being nonlinear everywhere, do not lead to improvement on four out of five measures when they replace min-out.
It also has difficulty in explaining why max-out only has a minor impact on robustness while min-out increases performance as much as five times (from 0.03 to 0.16 for BIM).

Training with independent cross entropy loss has a mixed effect on robustness. Adversarial perturbation is an inherently local phenomenon. The effect of activation suppression in distant areas to the decision boundary around an example is a poorly-understood area and beyond the scope of the current paper.

\subsection{Robustness on CIFAR-10}

\begin{table*}[t]
    \centering
    \small
\begin{tabular}{rlccccc}
\toprule
 & \multirow{2}{*}{Model} & Clean & 
 FGM & FGM & C\&W & BIM \\
 & & & $\epsilon_{L_\infty}=0.3$ & $\epsilon_{L_2}=2$ & & $\epsilon=0.3$ \\
\midrule
0 &             Baseline & 0.75\hspace{0.5em} &           0.08\hspace{0.5em} &           0.43\hspace{0.5em} &       0.01\hspace{0.5em} &        0.01\hspace{0.5em} \\
1 &              + ReLog ($\beta=1$) &     0.75\hspace{0.5em} &       0.16$^*$ &           \textbf{0.45}\hspace{0.5em} &       0.03$^*$ &        0.01\hspace{0.5em} \\
2 &       + MaxOut ($k=4$) & 0.65$^*$ &           0.12$^*$ &           0.31$^*$ &       0.02$^*$ &        0.01$^*$ \\
3 &       + MinOut ($k=2$) & 0.70$^*$ &           \textbf{0.24}$^*$ &           0.37$^*$ &       \textbf{0.04}$^*$ &        0.01$^*$ \\
4 &         + Elliptical ($\alpha=0.5$) &   0.68$^*$ &         0.12$^*$ &           0.35\hspace{0.5em} &       0.02$^*$ &        0.01\hspace{0.5em} \\
5 &        + MaxFit ($L_1$) &  0.76$^*$  &         0.10$^*$ &           0.41$^*$ &       0.02\hspace{0.5em} &        0.00$^*$ \\
6 &       + BCE training &   0.78\hspace{0.5em} &         0.12$^*$ &           0.42\hspace{0.5em} &       \textbf{0.04}$^*$ &        0.01$^*$ \\
7 &  + Negative examples &     0.66\hspace{0.5em} &       0.13\hspace{0.5em} &           0.32$^*$ &       \textbf{0.04}\hspace{0.5em} &        \textbf{0.03}$^*$ \\
\bottomrule
\end{tabular}
    \caption{Accuracy of models on CIFAR-10 on different attacks. BIM: iterations=5, C\&W: iterations=50. Elliptical units are applied to the first hidden layer only. An asterisk ($^*$) signifies that the result is statistically significantly different from the previous one on the same column. }
    \label{tab:accuracy-cifar10}
\end{table*}

The improvement of robustness on CIFAR-10 is more modest (see Table~\ref{tab:accuracy-cifar10}).
Modified models work the best against FGM ($L_\infty$) and C\&W attacks for which they are always better than the baseline. However, they are often only as good as the baseline for BIM and sometimes underperform the baseline for FGM ($L_2$).
Improvement coming from ReLog is more consistent than other traits as the corresponding model outperforms the baseline in two measures and maintain statistically equivalent performance on the rest.

We again observe that min-out is superior than both max-out and elliptical units, a property cannot be explained by the linearity hypothesis.
The results of BCE training mirrors what has been observed in MNIST experiments.

\subsection{Discussions}

\begin{figure}[t]
    \centering
    \includegraphics[width=0.25\textwidth]{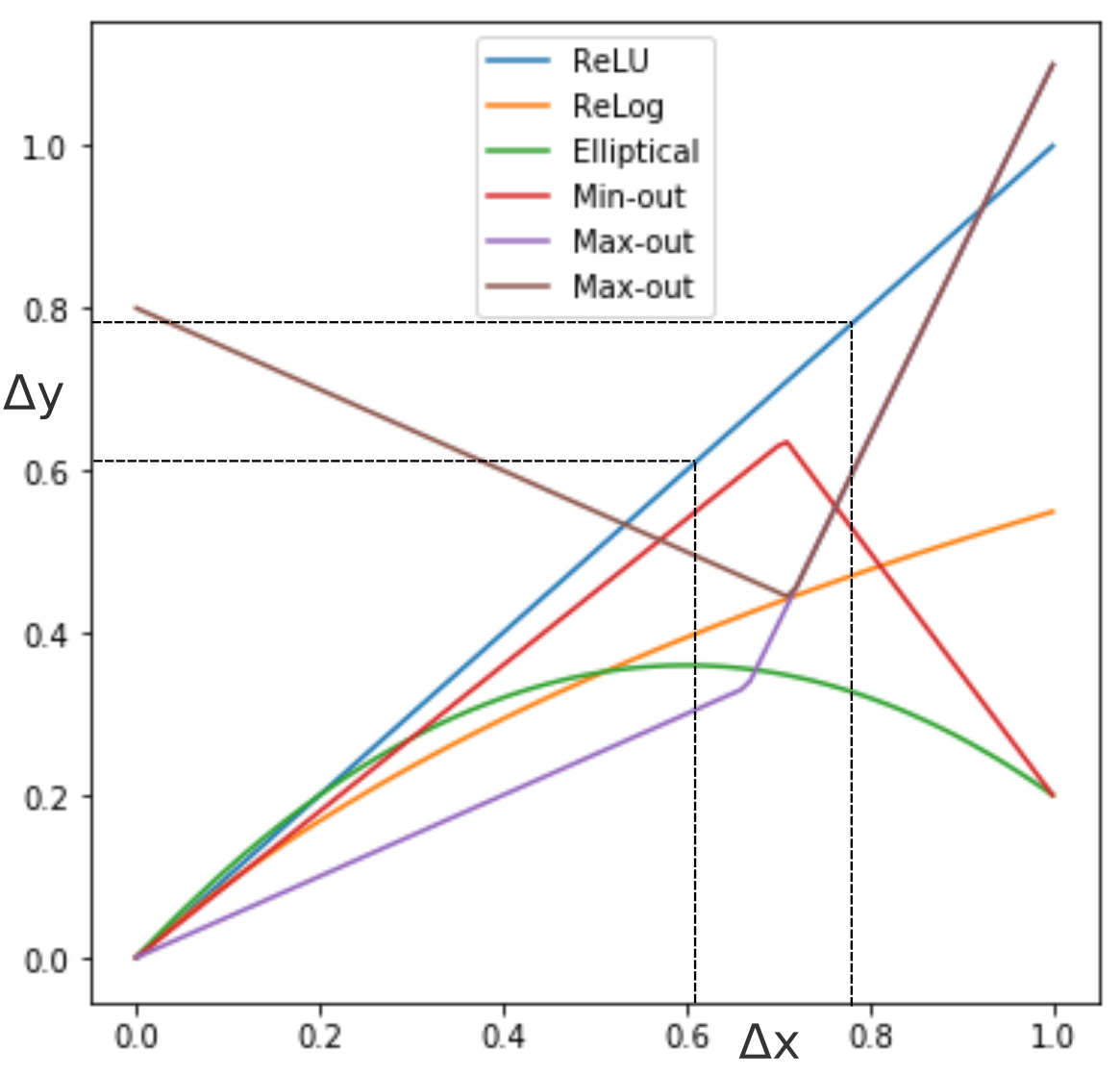}
    \caption{The change of output in response to a small perturbation of input: $\Delta_y = f(x+\Delta_x) - f(x)$ for different functions $f$.}
    \label{fig:activations}
\end{figure}{}

Given the experimental results, a revision to the linearity hypothesis is needed.
Hoffman et al.~\shortcite{Hoffman2019} recently showed that, theoretically and empirically, smaller norms of the input-output Jacobian matrix implies higher robustness.
This result explains why ReLog is consistently better than ReLU and adding elliptical units is advantageous.
However, it has difficulty in explaining min-out and max-out results because adding them do not change the locally linear behavior of a ReLU neural network.

To propose a unified framework that explains all observations, let us consider the input-output finite difference instead:
$\Delta_y = f(x+\Delta_x) - f(x)$
where $f$ is a neural network, $x \in \mathbb{R}^m, y=f(x) \in \mathbb{R}^n$ are its input and output, and $\Delta_x \in \mathbb{R}^m$ is a vector of a small norm: $|\Delta_x|_p \le \epsilon$.
Figure~\ref{fig:activations} depicts a special case where $m=n=1$ or, equivalently, where we consider only one neuron and one linear direction in the input space.

From the graph, it can be inferred that ReLog and elliptical units are inherently more robust than ReLU and the effect increases monotonically with curvature. Min-out significantly improves stability as long as the input lies close to the intersection of two hyperplanes whereas max-out can either increases or decreases stability. The inferior performance of elliptical units compared to min-out perhaps reflects a difficulty in training instead of an inherent property.

\section{Related Work}
\label{sec:related-work}


\textbf{Neural Network Design.}
Some elements of the architecture proposed here were introduced before in isolation. 
Training with negative examples were used in \cite{Bromley.denker93} to encourage the rejection of ``rubish class'' examples.
Maxout units were designed by \cite{goodfellow2013maxout.manual} and originally intended to replace rectified linear units instead of working together with them.
Parallel to this research, Fan and Wang~\shortcite{Fan2019.manual} succeeded in training quadratic units. Finally, Liu et al.~\shortcite{Liu19} proposed logarithmic activation functions but their formulations are slightly different from ours and lack the smooth transition from ReLU.
Different from all the work cited above, I combine various non-conventional design elements and show that they work together to make a neural network more local and robust.

\textbf{Adversarial Examples.}
Because of practical importance, the majority of the literature is concerned with alleviating adversarial examples. 
Most papers focus on improving conventional architectures through regularization \cite{Mustafa19} or injection of adversarial examples into training \cite{Kurakin2017,Madry2017}.
Results that link robustness to sparseness \cite{Guo2018.short} and margin \cite{Croce2018,Galloway2018} can be considered a special case of the input-output Jacobian approach taken by Hoffman et al.~\shortcite{Hoffman2019} which is closely related to my finite difference analysis.
Taghanaki et al.~\shortcite{Taghanaki2019} show that another locality-inducing function, the radial basis functions (RBFs), can reduce the success rate of attacks, as might be expected from the analysis in the previous section.

Because experimental settings vary a lot from one paper to another, it is often hard to compare reported results across papers.
To compare against Taghanaki et al.~\shortcite{Taghanaki2019}, I have set up attacker models to the same settings.
Their RBF models significantly outperform ones in this paper on MNIST but this result does not affect my conclusions.




\textbf{Fooling Examples.}
Little work has been done in this area. A rare find is Ghosh et al.~\shortcite{Ghosh2019} which uses an autoencoder with a mixture of Gaussian prior to detect and reject fooling examples.

\section{Conclusions}
\label{sec:conclusions}

This paper brings in ideas from fuzzy logic to improve deep neural networks. 
Preliminary results show that the proposed models are more well-behaving on noise patterns and more robust against adversarial examples. 
Analyses confirm that both the proposed architecture and training procedure contribute to performance on noise patterns while the architectural modifications improve robustness by enabling multiple disconnected regions and increasing stability in each one.

Perhaps most importantly, the current paper hints at what deep neural networks might be capable of: being both local and deep, and combining the learnability and expressiveness of connectionism and the validity of logic.
Exciting lines of future research can be foreseen:  expanding and generalizing the proposed architectures, improving training procedures, improving calibration on noise patterns and other stimuli, and studying the interaction between local and distributed representations.

\bibliography{Mendeley-Minh,manual}
\bibliographystyle{named}

\end{document}